\documentclass{IOS-Book-Article}

\usepackage{amsmath}
\usepackage{mathtools}
\usepackage{mathptmx}
\usepackage{url}
\usepackage{hyperref}
\usepackage{graphicx}
\usepackage[noabbrev, capitalize]{cleveref}
\usepackage{booktabs}
\usepackage{subcaption}
\DeclarePairedDelimiter{\abs}{\lvert}{\rvert}

\def\hb{\hbox to 10.7 cm{}}
\begin{document}

\pagestyle{headings}
\def\thepage{}

\begin{frontmatter}              

    \title{Towards an Argument Mining Pipeline Transforming Texts to Argument Graphs}

    \markboth{}{September 2020\hb}

    \author{\fnms{Mirko} \snm{Lenz}%
    \thanks{Corresponding author. E-mail: \texttt{info@mirko-lenz.de}}},
    \author{\fnms{Premtim} \snm{Sahitaj}},
    \author{\fnms{Sean} \snm{Kallenberg}},
    \author{\fnms{Christopher} \snm{Coors}},
    \author{\fnms{Lorik} \snm{Dumani}},
    \author{\fnms{Ralf} \snm{Schenkel}}, and
    \author{\fnms{Ralph} \snm{Bergmann}}

    \runningauthor{Lenz et al.}
    \address{Trier University, Trier, Germany}

    \begin{abstract}
        This paper targets the automated extraction of components of argumentative information and their relations from natural language text. Moreover, we address a current lack of systems to provide complete argumentative structure from arbitrary natural language text for general usage. We present an argument mining pipeline as a universally applicable approach for transforming German and English language texts to graph-based argument representations. We also introduce new methods for evaluating the results based on existing benchmark argument structures. Our results show that the generated argument graphs can be beneficial to detect new connections between different statements of an argumentative text. Our pipeline implementation is publicly available on GitHub.
    \end{abstract}

\end{frontmatter}
\markboth{September 2020\hb}{September 2020\hb}

\section{Introduction}\label{sec:introduction}

Argumentation plays an integral role in many aspects of daily human interaction. Its influence can be observed in different areas ranging from organizational decision-making to investigative journalism. People use arguments to form opinions, discuss ideas or change the views of others.
Many resources dealing with argumentation are available, but the content is mostly unstructured.
Due to the current capabilities of modern computation devices, Computational Argumentation (CA) is a field of increasing interest which is heavily investigated in the context of Natural Language Processing (NLP).
%
The ReCAP project~\cite{Bergmann2018} targets the realization of an Argumentation Machine~\cite{reed2003} that primarily operates on the knowledge level by enabling argument-based reasoning. Novel methods are developed that capture arguments in a robust and scalable manner and make argument information available to decision makers, journalists and researchers through contextualization, representation, and aggregation of argumentation.

While previous work~\cite{Levy2018, Stab2014, Florou2013, Goudas2014} has focused more on individual tasks such as claim detection~\cite{Levy2014, Lippi2016}, this paper targets the automated extraction of components of argumentative information and their relations from natural language text. We address a gap in the field of argument mining where end-to-end pipelines that generate complex argument structures for CA are not prevalent. We present an argument mining pipeline that provides an universally applicable approach for transforming German and English language texts to graph-based representations~\cite{Craven2016ArgumentGraphsAssumptionbased}, which can then be used to examine the argumentative structure both computationally and graphically. We also introduce new methods for evaluating the results based on existing benchmark argument structures and present a new argument graph dataset, which can be shared on request from the authors.

Next, \cref{sec:foundations} serves with necessary foundations to perform argument mining of graph structures and highlight relevant work in the field of CA. \Cref{sec:main} describes our proposed pipeline, while \cref{sec:evaluation} is concerned with a systematic evaluation. \Cref{sec:conclusion} concludes the paper and provides an outlook to further research in this field.

\section{Foundations and Related Work}\label{sec:foundations}

\begin{figure}[tb]
    \centering
    \begin{subfigure}[b]{.35\linewidth}
        \footnotesize
        \emph{Not all practices and approaches may have been proven in clinical trials,}
        so it makes no sense that
        \textbf{health insurance companies should naturally cover alternative medical treatments.}
        \vspace{1.5em}
        \caption{ADUs where the claim is bold and the premise underlined.}
    \end{subfigure}
    \begin{minipage}[b]{.15\linewidth}
        \centering
        \large
        \(\implies\)
        \vspace{22.5mm}
    \end{minipage}
    \begin{subfigure}[b]{.35\linewidth}
        \includegraphics[width=\linewidth]{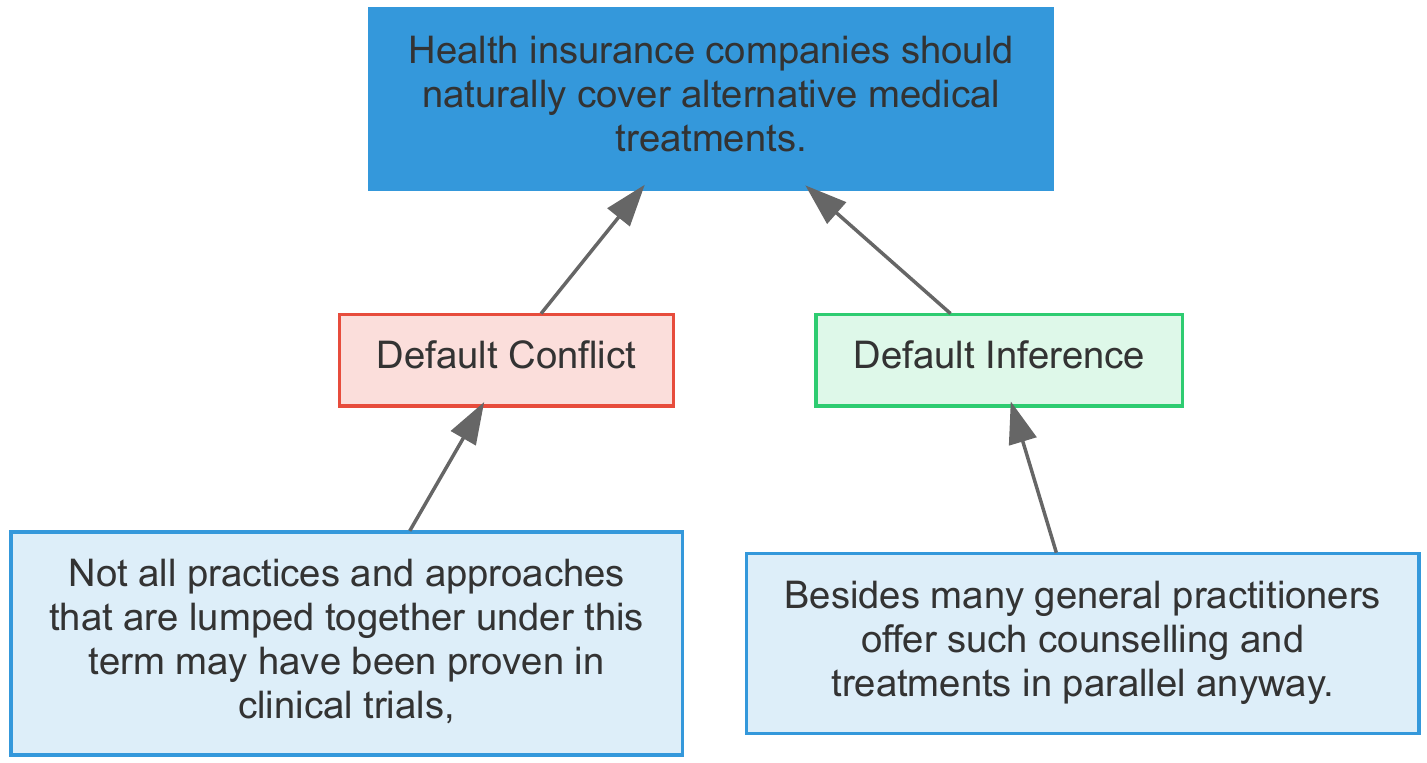}
        \caption{Resulting argument graph with highlighted major claim.}
    \end{subfigure}
    \caption{Example of ADUs (a) and the corresponding argument graph (b) from the microtexts corpus~\cite{Peldszus2015}.}\label{fig:adus-graph}
\end{figure}

Argumentation, in a formal way, is described as a set of arguments in texts. An argument consists of a claim and at least one premise. A \emph{claim}, which represents a controversial statement, can either be supported or attacked by one or multiple \emph{premises}, which provide the actual evidence to a claim. Further, the \emph{major claim} is defined as the claim that describes the key concept in an argumentative text~\cite{Stab2014}. Major claims, claims and premises are considered \emph{Argumentative Discourse Units} (ADUs) and represent the components of argumentation~\cite{Stab2014}. Additionally, we can represent the stance between two ADUs as a supporting or attacking directed relation.
An \emph{argument graph} describes a structured representation of argumentative text~\cite{DBLP:conf/lrec/StedeAPAP16}.
We use a variant of the well-known Argument Interchange Format (AIF)~\cite{Chesnevar2006}, extended to support the explicit annotation of a major claim \(M\)~\cite{DBLP:conf/iccbr/LenzOSB19}.
Claims, premises and the major claim are represented as \emph{information nodes} (I-nodes) while relations between them are represented by \emph{scheme nodes} (S-nodes).
The set of nodes \(V = I \cup S\) is composed of I- and S-nodes.
The supporting or attacking relations are encoded in a set of edges \(E \subseteq V \times V\).
Based on this, we define an argument graph \(G\) as the triple \(G = (V,E,M)\).
To illustrate the connection between ADUs in a text and I-nodes in a graph, \Cref{fig:adus-graph} visualizes a potential argument mining result where natural language text (a) was transformed into an argument graph (b).

We aim at addressing the research gap of a general-use end-to-end pipeline for the German and English languages by following and extending the approaches of related work in the field of argumentation. Cabrio and Villata~\cite{Cabrio2018} propose argumentation mining from a data-driven perspective in a survey. They further define the central stages of an argument mining framework: argument extraction and relation prediction.
Stab and Gurevych~\cite{Stab2014} present an approach for identifying ADUs in persuasive essays with further classification into major claim, claims and premises through a multiclass classifier which was trained on a set of structural, lexical, syntactic and contextual features~\cite{Lippi2016,Levy2014,Florou2013}. Goudas et al.~\cite{Goudas2014}, Wachsmuth et al.~\cite{DBLP:conf/coling/WachsmuthKS16}, and Stab et al.~\cite{DBLP:conf/emnlp/StabMSRG18} simplified the segmentation of natural language text into ADUs by considering textual boundaries on the sentence level.
Argument extraction is the prerequisite for the following central stage of relation prediction. Many researchers formulate a binary classification problem to distinguish between supporting and attacking relations, which reduces the inherent complexity of the task~\cite{Stab2017}. The predicted relations are then used to construct an argument graph from the extracted ADUs~\cite{Cabrio2018}. To the best of our knowledge, only Stab and Gurevych~\cite{Stab2014} addressed a method to link claims with premises within the same paragraph in an argumentative text.
Nguyen and Litman~\cite{DBLP:conf/aaai/NguyenL18} developed an argument mining system for the purpose of automated essay scoring. Here, a full pipeline was introduced including the identification of relevant ADUs, the classification of ADUs into major claim, claims and premises as well as the classification of relations between these ADUs. This system provides a complete but specialized end-to-end argument mining system only for the application on the persuasive essay domain.

For training and evaluating argument mining methods, there exists a diverse selection of corpora.
Stab et al.~\cite{stab-gurevych-2017-parsing,Eger:2018:Coling} provide a corpus with 402 annotated persuasive essays---in the following called PE.
Each ADU is labeled with ``Premise'', ``Claim'', or ``Major Claim'' and relationships between them are defined.
It consists of \(5,740\) I-nodes, \(5,338\) S-nodes and \(10,676\) edges.
Another corpus has been developed by the ReCAP project~\cite{Dumani2020}, consisting of 100 argument graphs.
The original texts dealing with educational issues in Germany are very diverse, ranging from press releases to newspaper articles.
There are \(2,533\) I-nodes, \(2,281\) S-nodes and \(4,838\) edges available.

\section{Argument Mining Pipeline}\label{sec:main}

The pipeline introduced by Nguyen and Litman~\cite{DBLP:conf/aaai/NguyenL18} is used as the basis of our proposed architecture and extended by a novel graph construction process.
Our pipeline is designed in a modular way where each step describes an individual and interchangeable module.
We start by addressing (1) argument extraction where input text is segmented and potential argumentative units are identified. Afterwards, task of (2) relationship type classification is considered, where extracted units are classified to determine their stance.
After (3) detecting the major claim, (4) an argument graph is constructed.

\subsection{Argument Extraction}

As a first step, the input text is segmented into sentences~\cite{Stab2014,Dumani2020}.
Then, multiple types of features are extracted, derived from Stab and Gurevych~\cite{stab-gurevych-2017-parsing} as well as Lippi et al.~\cite{Lippi2015}, depicted in \cref{tab:features}.
The basis of the entire approach is the correct identification of ADUs.
Based on these features, the sentences are classified into argumentative and non-argumentative units.
The ADUs are then further categorized into claims and premises using a separate classifier.

\begin{table}[tb]
    \centering
    \caption{Linguistic features extracted during preprocessing.}\label{tab:features}
    \begin{tabular}{lp{8cm}}
        \toprule

        \textbf{Category} & \textbf{Features}                                                                                                \\
        \midrule

        Structural        & Punctuation, sentence length and position.                                                                        \\
        Indicators        & Claim-premise and first-person indicators.                                                                                 \\
        Syntactic         & Depth of constituency parse trees, presence of modal verbs, number of grammatical productions in the parse tree. \\
        Embeddings        & GloVe sentence embeddings (arithmetic mean of its word vectors).                        \\
        \bottomrule
    \end{tabular}
\end{table}

\subsection{Relationship Type Classification}\label{sec:relation-classification}

To be able to construct an argument graph from a natural language text it is necessary to consider the task of textual entailment. Here, we assign the relation type between the identified ADUs~\cite{cabrio-villata-2012-combining}.
The direction of the inference is only considered from premise to claim.
Due to the complexity of considering a multi-class stance problem and the lack of training data of more sophisticated argument schemes (e.g.,  Walton et al.~\cite{Walton2008}), we only train a model to classify attacking and supporting relations.
Embeddings are used as the only feature for this task to focus on semantic information.
Based on the model's metadata, we detect indifferent results (i.e., results having a classification probability below a configurable threshold).
In this case, the type support is used.

\subsection{Major Claim Detection}
A very crucial step in the graph generation is locating the major claim.
To the best of our knowledge there are neither pre-trained models nor sufficient training data available, as each text usually has only one major claim, regardless of its length.
This makes machine learning-based approaches infeasible.
The classifier provided by Stab et al.~\cite{Stab2014} is not usable as it condenses all classification steps into a single model, which does not fit our proposed pipeline.
Thus, three different heuristics are available:

\paragraph{First}
Here, the first claim based on the text position is chosen as the major claim.
This is done due to usual text structures where the main argument is often referred to in the introduction or headline (e.g., Dumani et al.~\cite{Dumani2020} report that 58 of 100 major claims occur in the first quarter of the text).

\paragraph{Centroid}
When treating the major claim as the core proposition of the text, we can assume that it should be very similar to all ADUs.
Thus, we can compute the centroid of all embeddings to estimate the core message.
The major claim is then defined as the ADU with the highest cosine similarity to the centroid.

\paragraph{Pairwise}
The cross product of all embeddings of the ADUs is used to compute pairwise cosine similarity scores.
The major claim is defined as having the highest average similarity to all other ADUs.
The rational for this technique is similar to \textsc{Centroid}.

\paragraph{Probability}
Again, a cross product of all ADUs is computed.
Based on the relationship classification (see \cref{sec:relation-classification}), the major claim is defined as having the highest average classification probability (except for neutral results).
In other words, we select the ADU where the model shows the highest certainty in all of its predicted relations.

\subsection{Graph Construction}
Having obtained all necessary argumentative information, we can now construct the graph.
To the best of our knowledge, there is no automatic procedure that links ADUs to complex graphs.
We propose three algorithms to address this task.
In all cases, the ADUs are used as I-nodes and the S-nodes between them are derived from the relationship type classification.
As a simplification, the major claim is always set as the root.

\paragraph{Flat Tree}
Our baseline approach is to connect all ADUs as I-nodes to the major claim using the predicted S-nodes, resulting in a two-layer graph.
While not suitable for complex texts, it may still provide sufficient results for smaller ones or single arguments.

\paragraph{ADU Position}
This technique makes use of typical argument compositions.
We assume that premises belonging to a claim are contained in the same paragraph and thus positioned in close proximity of the claim in the original text~\cite{Stab2014}.
In the first step, all claim I-nodes are connected to the major claim using the respective S-nodes.
Then, each a premise I-node is connected to the nearest claim via an S-node.
If no claim is detected, all premise I-nodes are connected directly to the major claim via S-nodes.
The resulting graph consists of at least two and at most three layers.

\paragraph{Pairwise Comparison}
This method leverages the class probabilities of the relationship type classification.
The basic idea is to draw an edge between ADUs whose relation probability is above a certain threshold.
First of all, tuples of ADUs \((a,b)\) are computed such that \(b\) has the highest relation probability among all possible connections of \(a\).
If multiple ADUs reach the same maximal value, the first one is chosen.
Then, a configurable lower bound (in our case 0.98) below this maximal probability is defined.
Each ADU that is related to the major claim with a score above the lower bound is connected as an I-node via a corresponding S-node.
If the major claim has no connections after this step, the ADU that first occurs in the text is used an an I-node and connected to the major claim.
Then, the remaining ADUs are connected iteratively (via S-nodes) to the I-node where their score is above the lower bound.
If there remain ADUs not used after a certain amount of repetitions, they are connected to the major claim using a support S-node.

\section{Experimental Evaluation}\label{sec:evaluation}
In this section we evaluate our end-to-end approach by assessing the resulting argument graph structures.
Moreover, we compare the correspondence of our automatically generated graph to a given benchmark graph.


\subsection{Hypotheses}
The following hypotheses, covering all aspects of the pipeline, will be tested in our evaluation.
\textbf{H1} Using sentences as an argumentative unit yields a robust approximation of the manually segmented ADUs.
\textbf{H2} Selecting the major claim using \textsc{First} will result in the best results as it reflects common argumentation patterns.
\textbf{H3} Using a threshold for the relationship type classification (i.e., a value above 0.5) will perform best as supporting arguments occur more often than attacking ones.
\textbf{H4} Using \textsc{ADU Position} to construct graphs will result in the best approximation of the benchmark data due to the use of the claim-premise information.
\textbf{H5} Providing the pipeline with predefined ADUs will result in graphs that better reflect the human annotation than end-to-end graphs.

\subsection{Experimental Setup and Datasets}
The implementation has been done in Python and is available on GitHub\footnote{\url{https://github.com/ReCAP-UTR/Argument-Graph-Mining}, licensed under Apache 2.0.}.
The software is able to produce and evaluate two independent graphs:
(1) An end-to-end procedure that generates an argument graph by processing free-form text,
and (2) a procedure that uses ADUs from the benchmark data directly, skipping the sentence segmentation, ADU classification and major claim detection.
Three datasets are used for the evaluation:
ReCAP, PE (see \cref{sec:foundations}) and a new one created for our tasks.
The ReCAP corpus contains fragments such as headlines and metadata that were removed manually from the input files.
We are using two version of the PE dataset.
PE\textsubscript{17} is based on Stab et al.~\cite{stab-gurevych-2017-parsing}, converted to our AIF-based graph format.
The length of the ADUs differs greatly and is not in line with our sentence-based segmentation.
PE\textsubscript{18} is based on Eger et al.~\cite{Eger:2018:Coling} and was transformed by us from word- to sentence-based labels to conform to our segmentation approach.
A major difference is that PE\textsubscript{17} has information about relations between ADUs (i.e., is available as argument graphs), while PE\textsubscript{18} only provides the ADUs themselves.
We also explored the open discourse platform Kialo\footnote{\url{https://www.kialo.com}} due to the availability of much larger argument graphs (i.e., having more nodes/edges).
We limited the vast amount of potential debates to the 589 ones in the popular collection (as of Jan.\ 2020), consisting of \(190,269\) I-nodes, \(189,680\) S-nodes and \(379,360\) edges.
The data is available in English on request from the authors and has been translated to German via DeepL\footnote{\url{https://www.deepl.com}}.

\subsection{Classification Models}\label{sec:classification-models}
Having detailed the basic evaluation setup, we need to train the models for the ADU, claim-premise and relationship type classifiers.
As classification model for the ADU and claim - premise classification we chose an ensemble stacking method build from a layer of a logistic regression, random forest and adaptive boosted decision tree~\cite{Schapire1999} as they were shown to perform well for those specific tasks~\cite{aker-etal-2017-works}. The first layer of classifiers adds their predictions as feature to the input features and passes them on to the final estimator which provides the output prediction. For the output layer we chose a extreme gradient boosted random forest~\cite{DBLP:journals/corr/ChenG16}.
The ADU model was trained using the PE\textsubscript{18} and ReCAP datasets in their respective native languages (i.e., German for ReCAP and English for PE\textsubscript{18}).
The native languages were used to mitigate eventual translation errors.
The claim-premise classifier was trained using the PE\textsubscript{18} dataset for both languages as it is the only one that differentiates between claims and premises while also using sentences as units.
To eliminate biases, a 90/10 train/test split has been performed before training, other ratios were rejected because of the small size of the ReCAP corpus.
The models were trained through a 5-fold stratified cross-validation on the training set and tuned through a random search.
The reported values are results from a single evaluation on the test set.

We observed that the ADU classification reached highly varying results between the two datasets. We assume that the limited quantity of training data in the ReCAP dataset is the main reason for the variation.
On the more than four times larger essay data we obtained an accuracy score of 0.80 which yields a strong indication of the model's generalization ability.
The claim-premise classification unfortunately did not meet expectations on neither the persuasive essays nor on the ReCAP data. We explain the difference in predictive power on both datasets due to the fact that the structure of the ReCAP dataset is too dissimilar to the PE\textsubscript{18} dataset on which the models are trained on.
In \cref{tab:aduresult} we report accuracy \(A\), precision \(P\), recall \(R\) and \(F_1\) values for the used models.
\begin{table}[tb]
    \centering
    \caption{Results of the ADU and claim-premise classification models.}\label{tab:aduresult}
    \begin{subtable}[t]{.49\linewidth}
        \centering
        \caption{ADU model.}
        \begin{tabular}{lcccc}
            \toprule

            \textbf{Language} & \(A\) & \(P\) & \(R\) & \(F_1\) \\
            \midrule

            English (PE\textsubscript{18})      & 0.80  & 0.80  & 1.0   & 0.89    \\
            German (ReCAP)    & 0.54  & 0.52  & 0.66  & 0.58    \\
            \bottomrule
        \end{tabular}
    \end{subtable}
    \begin{subtable}[t]{.49\linewidth}
        \centering
        \caption{Claim-premise model.}
        \begin{tabular}{lcccc}
            \toprule

            \textbf{Language} & \(A\) & \(P\) & \(R\) & \(F_1\) \\
            \midrule

            English (PE\textsubscript{18})      & 0.52  & 0.52  & 0.68  & 0.59    \\
            German (PE\textsubscript{18})       & 0.76  & 0.73  & 0.13  & 0.22    \\
            \bottomrule
        \end{tabular}
    \end{subtable}
\end{table}

The training of the relationship type model has been performed using the Kialo dataset due to the large number of available relations.
The data was transformed into triples \(x, y, z\) with \(x\), \(z\) being I-nodes \(y\) being the S-node connected via edges to them.
We used models of type logistic regression (LG), k-nearest neighbors (KNN), random forest (RF) and gradient boosting (XGB) for the German and English languages.
The triples were split into 70\% training and 30\% testing data and achieved the results reported in \cref{tab:releval}.
XGB achieved the highest accuracy for both languages.

\begin{table}[tb]
    \centering
    \caption{Accuracy of the relation type classification models.}\label{tab:releval}
    \begin{tabular}{lcccc}
        \toprule

        \textbf{Language} & \textbf{LG} & \textbf{KNN} & \textbf{RF} & \textbf{XGB} \\
        \midrule

        English (Kialo)   & 0.6717      & 0.6005       & 0.6530      & 0.6783       \\
        German (Kialo)    & 0.6638      & 0.5851       & 0.6446      & 0.6677       \\
        \bottomrule
    \end{tabular}
\end{table}

\subsection{Argument Graph Metrics}
To assess the quality of the entire pipeline as well as its individual steps, multiple metrics are needed.
We are not aware of existing measures that enable the verification of our hypotheses, thus the following section introduces a novel approach.
For each element in the benchmark graph (i.e., I-nodes, S-nodes, major claim and edges), the corresponding item in the generated graph is determined to compute an agreement.

To compare the ADU segmentation, we need a mapping between the I-nodes of the benchmark graph \(G_b\) and the generated graph \(G_g\).
It is based on the Levenshtein distance~\cite{Levenshtein1966BinaryCodesCapable} \(\operatorname{dist}(u_b,v_g)\) between the benchmark I-node \(u_b\) and the generated I-node \(v_g\) and the derived similarity \(\operatorname{sim}(u_b, v_g) = 1 - (\operatorname{dist}(u_b,v_v) / \max\{\abs{u_b}, \abs{v_g}\})\).
The mapping \(m \colon u_b \mapsto v_g\) assigns each I-node of the benchmark graph an I-node of the generated graph s.t.\ their similarity is higher than any other combination of I-nodes.
A node in the generated graph cannot be mapped to more than one node in the benchmark graph.
In the case that two generated nodes have the same similarity to the benchmark node, the first one is selected.
If the ADU segmentation between the benchmark and generated graph differs (e.g., two sentences per ADU in the benchmark and one in the generated graph), the benchmark node is mapped to the generated node having the highest similarity while ignoring the other nodes.
The \emph{I-nodes agreement} \(\mathcal{I}\) is defined by the weighted arithmetic mean of the similarity between the benchmark I-nodes and their respective mappings.
The \emph{major claim agreement} \(\mathcal{M}\) is specified as a binary metric that is 1 if the major claims are mapped or there is none defined in the benchmark and 0 otherwise.

For the evaluation of S-nodes, we need to consider the surrounding I-nodes, because S-nodes do not contain textual content that could be used for similarity assessments.
We compute all combinations of connections of the benchmark S-node \(\operatorname{in}(u_b) \times \operatorname{out}(u_b)\) and determine individual tuples based on their respective mappings as \((m(\operatorname{in}), m(\operatorname{out}))\).
Using this information, it is possible to compare the benchmark S-node with the information provided by the relationship type classification.
The \emph{S-node agreement} \(\mathcal{S}\) is then defined as the number of correctly classified relationships divided by the total number of tuples.

As a last step, the edges need to be considered as well.
As they do not contain textual information, we look at the triple \((x, y, z)\) where \(x\) and \(z\) represent I-nodes and \(y\) an S-node.
Thus, we always look at two edges at a time.
The two edges in the benchmark graph are mapped to their counterparts in the generated graph if they connect the same I-nodes (as determined by the mapping \(m\)).
The direction of the edges is not relevant for this metric.
The S-node \(y\) is ignored deliberately to mitigate potential errors during earlier tasks.
The \emph{edges agreement} \(\mathcal{E}\) is then determined by dividing the number of mapped edges by the total number of available edges.
This metric has to be treated with caution, because the generation of argument graphs is highly subjective.
Even when done by trained professionals, the inter-annotator agreement is relatively low~\cite{Dumani2020}.

Lastly, the \emph{computation time} \(T\) is measured\footnote{On a 2019 MacBook Pro with a 2.3 GHz 8-core processor}.
We ignore the program initialization (i.e., loading data from disk to memory) and only consider the relevant processing time.

\subsection{Results and Discussion}
We will now present the evaluation of the pipeline using the test splits of the ReCAP dataset in German and both PE corpora in English.
After an analysis of the aggregated scores, an exemplary case will be discussed for each corpus.
The ReCAP corpus contains heterogeneous sets of text structures---ranging from newspaper articles to political proposals---which may cause worse results compared to the more uniform PE datasets.

\subsubsection{German ReCAP Corpus}
The test set for the ReCAP corpus contains ten texts with benchmark graphs.
We get an \emph{I-node} agreement \(\mathcal{I}=0.461\) for all possible combinations of parameters.
An in-depth look reveals that in most cases, there were fewer, but larger ADUs in the generated graph compared to the benchmark.
This stands in contrast to the fact that the average ADU length in the ReCAP corpus is 1.1, indicating mismatches in the definition of a sentence, for example due to punctuation.
It also contradicts \textbf{H1}.
\Cref{tab:eval-mc} shows the results of the three \emph{major claim} detection approaches.
They are very similar, differing only in one case (as we have exactly one major claim per text).
The two best methods \textsc{Centroid} and \textsc{Pairwise} predicted exactly the same major claims.
As \textsc{First} performed worst here, \textbf{H2} might be rejected.
All thresholds for the \emph{relationship type} classification are depicted in \cref{tab:eval-snodes}.
The best result can be obtained using 1.0 (i.e., the classifier always predicts support), which means that almost all of the relations in the benchmarks are of the type support.
With such a skewed distribution, this corpus may not be suitable to assess \textbf{H3}, thus we will postpone it to PE.
When comparing end-to-end with preset ADUs, we observe that the latter one delivers slightly worse performance with all thresholds above 0.6.
This could be caused by the smaller preset ADUs which provide less contextual information for the classifier.
This stands in slight contrast to \textbf{H5}.
Lastly, \cref{tab:eval-edges} shows the three \emph{graph construction} methods.
As the scores depend on the major claim method, we used the best approach (i.e., \textsc{Centroid/Pairwise}) for the end-to-end graph.
The algorithm \textsc{Flat Tree} delivered the best results across the board, contradicting \textbf{H4}.
As expected, the scores themselves are very low, especially for the end-to-end graph, making manual examination of individual edges necessary.
When comparing the end-to-end graph with the one using preset ADUs, we notice a major increase in the agreement score.
Using the best method, almost half of the edges were connected correctly, providing support for \textbf{H5}.
This is in large part caused by using the correct major claim as the root node.

\begin{table}[tb]
    \centering
    \caption{Aggregated results of the evaluation using the ReCAP corpus.}\label{tab:eval-recap}
    \begin{subtable}[t]{.24\linewidth}
        \centering
        \caption{Major claim methods.}\label{tab:eval-mc}
        \begin{tabular}{lc}
            \toprule

            \textbf{Method}      & \(\mathcal{M}\) \\
            \midrule

            \textsc{Centroid}    & \textbf{.200}   \\
            \textsc{First}       & .100            \\
            \textsc{Pairwise}    & \textbf{.200}   \\
            \textsc{Probability} & .100            \\

            \bottomrule
        \end{tabular}
    \end{subtable}
    \hfill
    \begin{subtable}[t]{.35\linewidth}
        \centering
        \caption{Relationship type thresholds.}\label{tab:eval-snodes}
        \begin{tabular}{ccc}
            \toprule

            \textbf{Threshold} & \(\mathcal{S}_{\text{e2e}}\) & \(\mathcal{S}_{\text{preset}}\) \\
            \midrule

            0.5                & .460                         & .514                            \\
            0.6                & .609                         & .699                            \\
            0.7                & .845                         & .792                            \\
            0.8                & .910                         & .884                            \\
            0.9                & .927                         & .898                            \\
            1.0                & \textbf{.937}                & \textbf{.902}                   \\

            \bottomrule
        \end{tabular}
    \end{subtable}
    \hfill
    \begin{subtable}[t]{.35\linewidth}
        \centering
        \caption{Graph construction methods (centroid major claim for end-to-end graph).}\label{tab:eval-edges}
        \begin{tabular}{lcc}
            \toprule

            \textbf{Method}         & \(\mathcal{E}_{\text{e2e}}\) & \(\mathcal{E}_{\text{preset}}\) \\
            \midrule

            \textsc{ADU Position}   & .064                         & .166                            \\
            \textsc{Flat Tree}      & \textbf{.095}                & \textbf{.449}                   \\
            \textsc{Pairwise Comp.} & .054                         & .296                            \\

            \bottomrule
        \end{tabular}
    \end{subtable}
\end{table}

\begin{figure}[tb]
    \includegraphics[width=.7\linewidth]{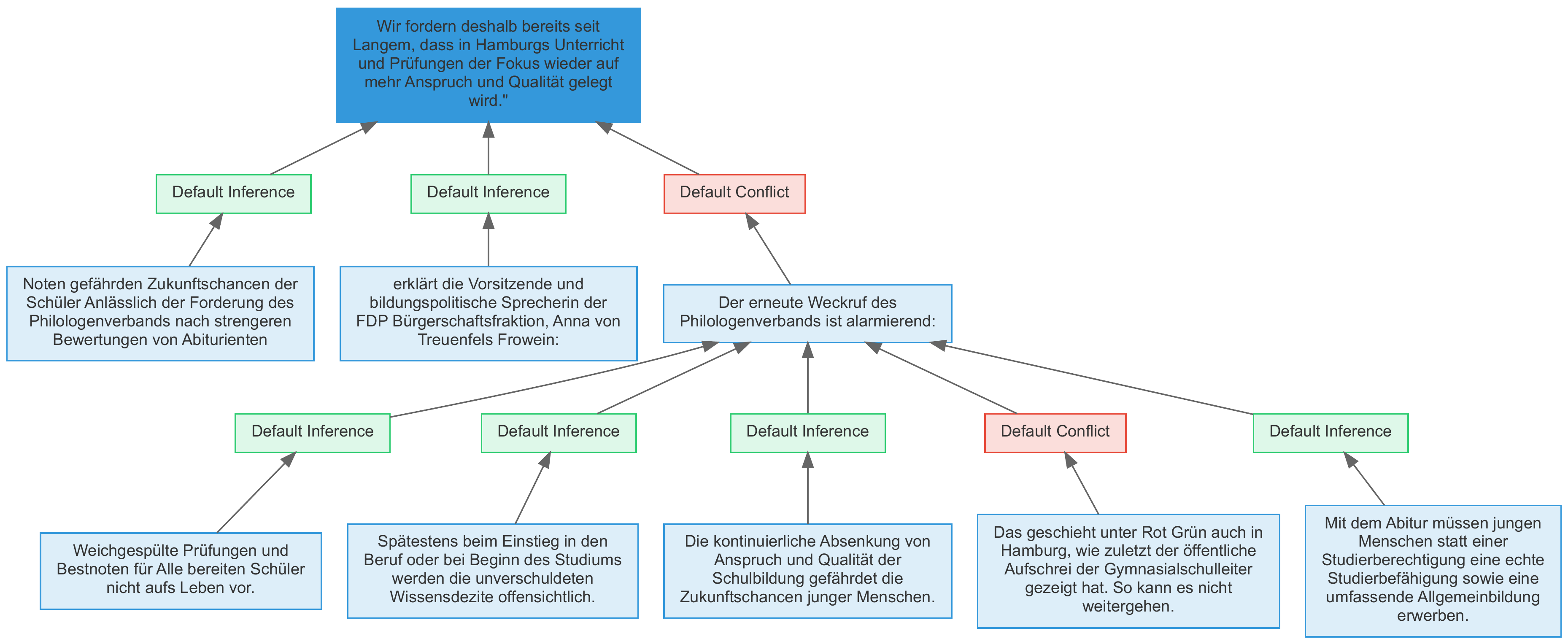}
    \caption{End-to-end graph from the ReCAP corpus generated by using \textsc{Centroid} major claim, a relationship type threshold of 0.6 and \textsc{ADU Position} to construct the graph, computed in 0.5 seconds.}\label{fig:eval-recap}
\end{figure}

In the following, an exemplary end-to-end graph based on the ReCAP corpus---depicted in \cref{fig:eval-recap}---is compared to its benchmark.
The automatically segmented ADUs are a bit longer than the benchmark ones, but still give a relatively good I-node agreement \(\mathcal{I} = 0.710\).
The major claims are nearly identical and only differ in the segmentation of the corresponding ADU, giving the agreement \(\mathcal{M}=1.0\).
The S-node matchings also obtain a high agreement \(\mathcal{S} = 0.667\).
In the benchmark graph, all relations are of the type support, while there are two attack ones in the generated graph.
Despite having a similar graphical structure, the edges themselves do not match well (i.e., having the agreement score \(\mathcal{E}=0.125\)).
They differ in the number of claims/premises as well as the node the premises are connected to.
However, an expert without knowledge about the benchmark graph assessed the chosen edges as reasonable.
Obviously, this is a subjective assessment that cannot be quantified.
The graph with preset ADUs shows better scores: \(\mathcal{S}=0.714\) and \(\mathcal{E}=0.429\).
This can mainly be attributed to the perfect segmentation.

\subsubsection{English PE Corpus}
For the following evaluation, the test split (see \cref{sec:classification-models}) of the PE corpus is used, consisting of 40 cases.
The results of PE\textsubscript{17} are very similar to the findings of the ReCAP corpus.
The I-node agreement \(\mathcal{I}=0.622\) is higher than for the ReCAP graphs, providing support for \textbf{H1}.
\textsc{Centroid} and \textsc{Pairwise} performed best for identifying the major claim (\(\mathcal{M}=0.1\)), contradicting \textbf{H2}.
A threshold of 0.9 for the relationship type classification yields the highest agreements (\(\mathcal{S}_{\text{e2e}} = 0.936\) and \(\mathcal{S}_{\text{preset}} = 0.912\)).
Again, the S-node distribution is skewed, but as two different corpora show the same results, we can accept \textbf{H3} for certain corpora.
The edge agreement scores can be obtained using \textsc{ADU Position} for the end-to-end graph (\(\mathcal{E}_{\text{e2e}}=0.130\)) and \textsc{Flat Tree} for the graph with preset ADUs (\(\mathcal{E}_{\text{preset}}=0.274\)).
All graph construction methods performed similarly and the agreement is relatively low, thus \textbf{H4} needs to be rejected.
The use of preset ADUs gives a big advantage in the edge agreement with only a small decrease in the S-node agreement, leading to the final acceptance of \textbf{H5}.
Overall, the findings show the robustness of the proposed approach for varying input data.
The PE\textsubscript{18} dataset gives another perspective on the pipeline by using sentence-based segmentation.
The I-node agreement \(\mathcal{I} = 0.799\) shows a decent approximation of the segmentation, leading to the partial acceptance of \textbf{H1} for certain corpora (e.g., essays).
The \emph{major claim} agreement \(\mathcal{M}\) is 0.125 for \textsc{Centroid} and \textsc{Pairwise}, 0.175 for \textsc{Probability} and 0.250 for \textsc{First}.
As \textsc{First} was only best in this specific corpus and the values are low overall, we have to reject \textbf{H2}.

\begin{figure}[tb]
    \includegraphics[width=\linewidth]{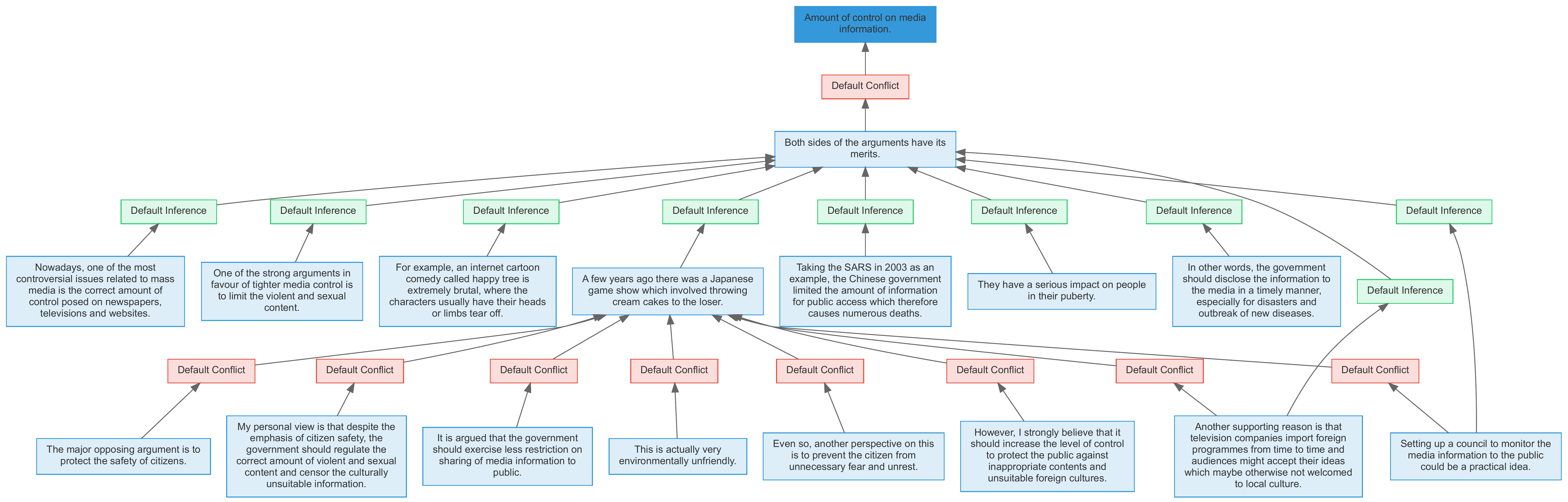}
    \caption{End-to-end graph from the PE\textsubscript{18} corpus generated by using \textsc{First} major claim, a relationship type threshold of 0.6 and and \textsc{Pairwise Comparison} to construct the graph, computed in 0.7 seconds.}\label{fig:eval-pe}
\end{figure}

\Cref{fig:eval-pe} shows an example of an end-to-end graph based on the PE\textsubscript{18} corpus.
The I-node agreement \(\mathcal{I}=0.765\) is slightly lower than average which is in this case caused by shorter ADUs compared to the benchmark data.
Some of the I-nodes are connected to multiple other nodes, showing the flexibility offered by the approach \textsc{Pairwise Comparison} (other methods only allow one outgoing edge per I-node).
However, the utility of such complex graphs for analyzing argumentative structures might be disputed.

\section{Conclusion and Future Work}\label{sec:conclusion}
In this work, we investigated new methods towards the automated mining of argument graphs from natural language texts for both English and German.
The pipeline successfully extends previous approaches~\cite{DBLP:conf/aaai/NguyenL18} by generating even complex graphs as the end product.
Our results show that there are great differences in the resulting graphs based on the type of input data.
For very homogeneous corpora such as PE, the agreement is very high, but in heterogeneous datasets such as ReCAP, the methods performed poorer.
When looking beyond the goal to approximate a human annotation as much as possible, the generated graphs might be very beneficial to detect new connections between single statements of an argumentative text.
Using multiple methods to construct different representations from a single text might also help in educating professional annotators by discussing the strengths and weaknesses of individual cases.

In future work we plan to provide a more flexible approach for segmenting a text into potential ADUs.
A limitation of the current evaluation procedure lies in the edge agreement, which could be tackled by providing multiple benchmark graphs to account for uncertainty.
As the ReCAP corpus makes use of detailed argumentation schemes~\cite{Walton2008}, the pipeline should be extended make use of them.
Finally, we will investigate the potential use of argument graphs for the task of measuring argument quality~\cite{DBLP:conf/eacl/WachsmuthSHPBHN17} in unstructured texts through the use of argument mining.

\paragraph*{Acknowledgments}
This work has been funded by the Deutsche Forschungsgemeinschaft (DFG) within the project \emph{ReCAP}, Grant Number 375342983 (2018--2020), as part of the Priority Program ``Robust Argumentation Machines (RATIO)'' (SPP-1999).
We would also like to thank \emph{DeepL} for providing free access to their translation API.

\bibliography{references}

\bibliographystyle{vancouver}
\end{document}